\documentclass[journal]{IEEEtran}

\usepackage[noadjust]{cite}
\usepackage{graphicx}
\usepackage{amsmath}
\usepackage{amssymb}
\usepackage{multirow}
\usepackage[letterpaper=true,colorlinks,bookmarks=false]{hyperref}
\usepackage{subfigure}
\usepackage[lined,boxed, ruled]{algorithm2e}

\begin{document}

\title{Joint Optic Disc and Cup Segmentation Based on Multi-label Deep Network and Polar Transformation}

\author{Huazhu~Fu, Jun~Cheng, Yanwu~Xu, Damon~Wing~Kee~Wong, Jiang Liu, and Xiaochun Cao
	\thanks{H.~Fu and D.~W.~K.~Wong are with Institute for Infocomm Research, Agency for Science, Technology and Research, Singapore 138632 (e-mail: huazhufu@gmail.com, wkwong@i2r.a-star.edu.sg).}
	\thanks{J.~Cheng is with Institute for Infocomm Research, Agency for Science, Technology and Research, Singapore 138632, and also with the Cixi Institute of Biomedical Engineering, Chinese Academy of Sciences, Zhejiang 315201, China (e-mail: sam.j.cheng@gmail.com)}
	\thanks{Y.~Xu is with Guangzhou Shiyuan Electronics Co., Ltd.~(CVTE), Guangzhou 510670, China (e-mail: xuyanwu@cvte.com).}
	\thanks{J.~Liu is with the Cixi Institute of Biomedical Engineering, Chinese Academy of Sciences, Zhejiang 315201, China (e-mail: jimmyliu@nimte.ac.cn).}
	\thanks{X.~Cao is with the State Key Laboratory of Information Security, Institute of Information Engineering, Chinese Academy of Sciences, Beijing 100093, China (e-mail: caoxiaochun@iie.ac.cn)}}


\maketitle

\begin{abstract}
	
Glaucoma is a chronic eye disease that leads to irreversible vision loss. The cup to disc ratio (CDR) plays an important role in the screening and diagnosis of glaucoma. Thus, the accurate and automatic segmentation of optic disc (OD) and optic cup (OC) from fundus images is a fundamental task. Most existing methods segment them separately, and rely on hand-crafted visual feature from fundus images. In this paper, we propose a deep learning architecture, named M-Net, which solves the OD and OC segmentation jointly in a one-stage multi-label system. The proposed M-Net mainly consists of multi-scale input layer, U-shape convolutional network, side-output layer, and multi-label loss function. The multi-scale input layer constructs an image pyramid to achieve multiple level receptive field sizes. The U-shape convolutional network is employed as the main body network structure to learn the rich hierarchical representation, while the side-output layer acts as an early classifier that produces a companion local prediction map for different scale layers. Finally, a multi-label loss function is proposed to generate the final segmentation map. For improving the segmentation performance further, we also introduce the polar transformation, which provides the representation of the original image in the polar coordinate system. The experiments show that our M-Net system achieves state-of-the-art OD and OC segmentation result on ORIGA dataset. Simultaneously, the proposed method also obtains the satisfactory glaucoma screening performances with calculated CDR value on both ORIGA and SCES datasets.
	
\end{abstract}

\begin{IEEEkeywords}
	Deep learning, optic disc segmentation, optic cup segmentation,  glaucoma screening, cup to disc ratio.
\end{IEEEkeywords}

\IEEEpeerreviewmaketitle

\section{Introduction}
 
Glaucoma is the second leading cause of blindness worldwide (only second to cataracts), as well as the foremost cause of irreversible blindness~\cite{Tham2014}. Since vision loss from glaucoma cannot be reversed, early screening and detection methods are essential to preserve vision and life quality. One major glaucoma screening technique is optic nerve head (ONH) assessment, which employs a binary classification to identify the glaucomatous and healthy subjects~\cite{Garway-Heath352}. However, the manual assessment by trained clinicians is time consuming and costly,  and not suitable for population screening. 

\begin{figure}[!t]
	\centering
	\includegraphics[width=1\linewidth]{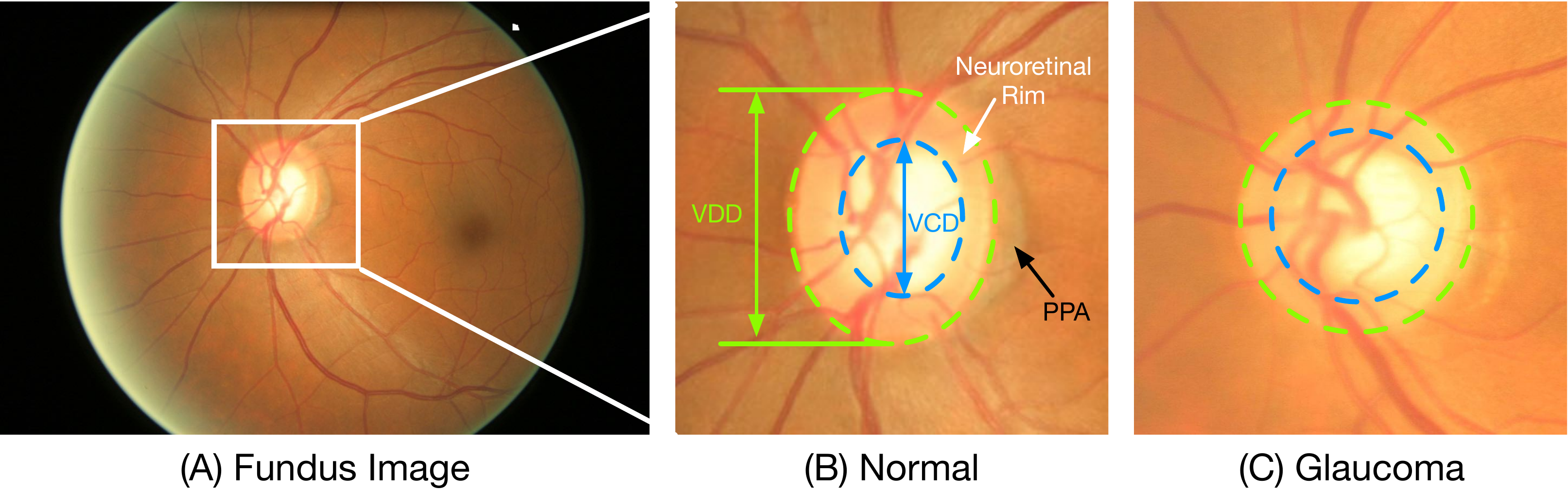}
	\caption{Structure of the optic nerve head. The region enclosed by the green dotted circle is the optic disc (OD); the central bright zone enclosed by the blue dotted circle is the optic cup (OC); and the region between them is the neuroretinal rim. The vertical cup to disc ratio (CDR) is calculated by the ratio of vertical cup diameter (VCD) to vertical disc diameter (VDD).  PPA: Peripapillary Atrophy.}
	\label{img-cover}
\end{figure}

For large-scale screening, automatic ONH assessment methods are needed. Some clinical measurements are proposed, such as the vertical cup to disc ratio (CDR)~\cite{Jonas2000}, rim to disc area ratio (RDAR), and disc diameter~\cite{HANCOXOD199959}.  In them, CDR is well accepted and commonly used by clinicians.   In color fundus image,  the optic disc (OD) appears as a bright yellowish elliptical region, and can be divided into two distinct zones:  a central bright zone as optic cup (OC) and a peripheral region as the neuroretinal rim, as shown in Fig.~\ref{img-cover}.  The CDR is calculated by the ratio of vertical cup diameter (VCD) to vertical disc diameter (VDD). In general, a larger CDR suggests a higher risk of glaucoma and vice versa.  Accurate segmentations of OD and OC are essential for CDR measurement. Some methods automatically measure the disc and cup from 3-D optical coherence tomography (OCT)~\cite{Lee2010,WuMenglin15,Fu2015TBME,Fu2017TMI}. However, OCT is not easily available due to its high cost, the fundus image is still referred to by most clinicians.
A number of works have been proposed to segment the OD and/or OC from the fundus image~\cite{Joshi2011,Cheng2013,Zheng2013miccai,Almazroa2015}. 
The main segmentation techniques include color and contrast thresholding, boundary detection, and region segmentation methods~\cite{Almazroa2015}. In these methods, the pixels or patches of fundus images are determined as background, disc and cup regions, through a learned classifier with various visual features. However, most existing methods are based on hand-crafted features (e.g., RGB color, texture, Gabor filter, and gradient), which lack sufficiently discriminative representations and are easily affected by pathological regions and low contrast quality. In addition, most methods segment the OD and OC separately, i.e., segmenting OD first, followed by the OC without considering the mutual relation of them. In this paper, we consider OD and OC together, and provide a one-stage framework based on deep learning technique.

Deep learning techniques have been recently demonstrated to yield highly discriminative representations that have aided in many computer vision tasks. For example, Convolutional Neural Networks (CNNs) have brought heightened performance in image classification~\cite{Krizhevsky2012} and segmentation~\cite{Long2017_FCN}. For retinal image, Gulshan~\textit{et al.} have demonstrated that the deep learning system could obtain a high sensitivity and specificity for detecting referable diabetic retinopathy~\cite{Gulshan2016}.  In fundus vessel segmentation, the deep learning systems~\cite{Fu2016ISBI,Fu2016,Maninis2016} also achieve the state-of-the-art performances. These successes have motivated our investigation of deep learning for disc and cup segmentation from fundus images. 

In our paper, we address OD and OC segmentation as a multi-label task and solve it  using a novel end-to-end deep network. The main contributions of our work include: 
\begin{enumerate}
	\item We propose a fully automatic method for joint OD and OC segmentation   using a multi-label deep network, named M-Net. Our M-Net is an end-to-end deep learning system, which contains a multi-scale U-shape convolutional network with the side-output layer to learn discriminative representations and produces segmentation probability map.
	\item For joint OD and OC segmentation, a multi-label loss function based on Dice coefficient is proposed, which deals well with the multi-label and imbalance data of pixel-wise segmentation for fundus image.
	\item Moreover, a polar transformation is utilized in our method to transfer the fundus image into a polar coordinate system, which introduces the advantages of spatial constraint, equivalent augmentation, and balancing cup proportion, and  improves the segmentation performance.
	\item At last, we evaluate the effectiveness and generalization capability of the proposed M-Net on ORIGA dataset. Our M-Net achieves state-of-the-art segmentation performance, with the average overlapping error of $0.07$ and $0.23$ for OD and OC segmentation, respectively. 
	\item Furthermore, the CDR is calculated based on segmented OD and OC for glaucoma screening. Our proposed method obtains highest performances with areas under curve (AUC) of $0.85$ and $0.90$ on ORIGA and SCES datasets.
\end{enumerate}

The remainders of this paper are organized as follows. We begin by reviewing techniques related to OD/OC segmentation in Section~\ref{sec-related}. The details of our system and its components are presented in Section~\ref{sec-method}. To verify the efficacy of our method, extensive experiments are conducted in Section~\ref{sec_exp}, and then we conclude with final remarks in Section~\ref{sec_conclusion}.

\section{Related Works}
\label{sec-related}

\textbf{Optic Disc Segmentation:}  The OD is the location where ganglion cell axons exit the eye to form the optic nerve through which visual information of the photo-receptors is transmitted to the brain. Earlier, the template based methods are proposed firstly to obtain the OD boundary. For example, Lowell \textit{et al.}~employed the active contour model~\cite{Lowell2004} to detect the contour based on image gradient. In~\cite{Aquino2010,Lu2011}, the Circular-based Transformation techniques are employed to obtain the OD boundary. In~\cite{Joshi2011}, the local texture features around each point of interest in multidimensional feature space are utilized to provide robustness against variations in OD region. Recently, the pixel classification based method is proposed to transfer the boundary detection problem into a pixel classification task, which obtains a satisfactory performance. Cheng \textit{et al.}~\cite{Cheng2013} utilizes superpixel classifier to segment the OD and OC, which exploits using various hand-crafted visual features at superpixel level to enhance the detection accuracy. In~\cite{Abra2007}, the disparity values extracted from stereo image pairs are introduced to distinguish the OD and background. However, reliance on hand-crafted features make these methods susceptible to low quality images and pathological regions. 

\textbf{Optic Cup Segmentation:} The OC is restricted to the region inside OD. Segmenting OC from fundus images is a more challenging task due to the low contrast  boundary. In~\cite{Wong2008}, an automatic OC segmentation algorithm based on a variational level set is proposed. Later, the  blood vessel kinks are found to be useful for OC segmentation~\cite{WongCup2009}, and a similar concept but named as vessel bend is utilized in~\cite{Joshi2011}. The main challenge in detecting kinks or vessel bending is that it is often affected by natural vessel bending that does not lie on the OC boundary. Moreover, the pixel classification based methods similar to OD segmentation~\cite{Cheng2013} is also introduced to OC segmentation. The various hand-crafted visual features (e.g., center surround statistics, color histogram, and low-rank superpixel representation) are employed in~\cite{Xu2011,Cheng2013,Xu2014} to represent the pixel/superpixel for OC segmentation. A common limitation of these algorithms is that they highly relied on hand-crafted visual feature, which are mainly based on contrast between the neuroretinal rim and the cup.  

\begin{figure*}[!t]
	\begin{center}
		\includegraphics[width=1\linewidth]{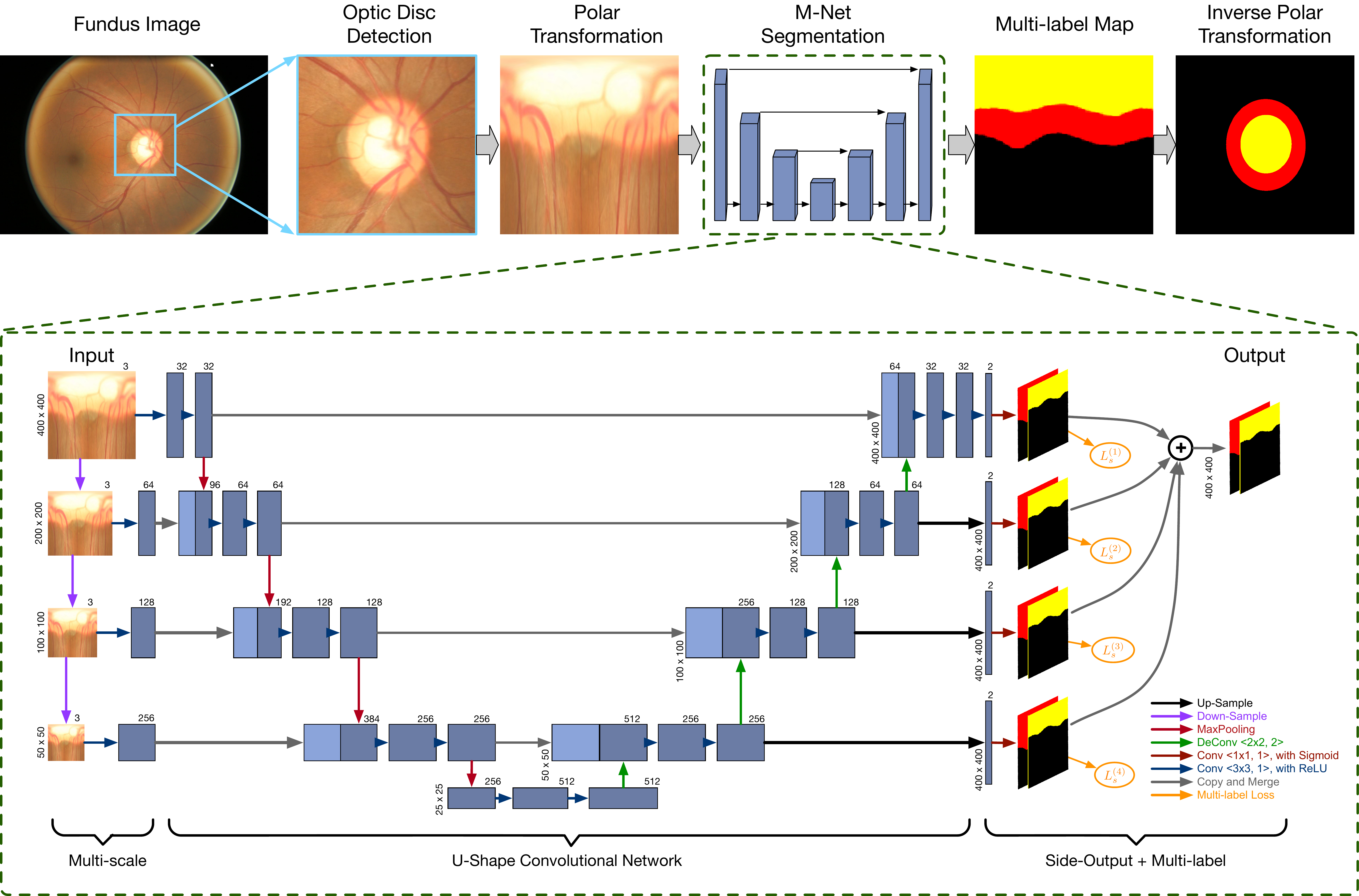}
		\caption{Illustration of our segmentation framework, which mainly includes fundus polar transformation and M-Net segmentation. Firstly, the optic disc is localized, and a polar transformation generates the representation of original fundus image in the polar coordinate system based on the detected disc center. Than our M-Net produces the multi-label prediction maps for disc and cup regions. Our M-Net architecture consists of multi-scale input layer, U-shape convolutional network, side-output layer, and multi-label loss function. The (De)Convolutional layer parameters are denoted as ``(De)Conv $<$kernel size, stride$>$".  Finally, the inverse polar transformation recovers the segmentation result back to the Cartesian coordinate. }
		\label{img-framework}
	\end{center}
\end{figure*}

\textbf{Joint OD and OC Segmentation:}  Most existing methods only focus on the single region segmentation (i.e., OC or OD). Especially, for the cup segmentation, the OD boundary could provide some useful prior informations, e.g., shape constraint and structure constraint~\cite{Xu2012b}. The works in~\cite{Joshi2011,Cheng2013} deal with the OD and OC by two separate stages with different features. Zheng~\textit{et~al.} integrated the OD and OC segmentation within a graph-cut framework~\cite{Zheng2013miccai}. However, they consider the OD and OC as two mutual labels, which means for any pixel in fundus, it can only belong to one label (i.e., background, OD, and OC). Moreover, the method~\cite{Zheng2013miccai} only employs color feature within a Gaussian Mixture Model to decide a posterior probability of the pixel, which makes it unsuitable for fundus image with low contrast. In~\cite{Sevastopolsky2017}, a modified U-Net deep network is introduced to segment the OD and OC. However, it still separates OD and OC segmentation in a sequential way. 
In~\cite{ZILLY201728}, an ensemble learning method is proposed to extract OC and OD based on the CNN architecture. An entropy sampling technique is used to select informative points, and then graph cut algorithm is employed to obtain the final segmentation result. However this multiple step deep system limits its effectiveness in the training phase.

\section{Proposed Method}
\label{sec-method}

Fig.~\ref{img-framework} illustrates the overall flowchart of our OD and OC segmentation method, which contains M-Net deep network and the fundus image polar transformation. In our method, we firstly localize the disc center  by using the existing automatic disc detection method~\cite{XU20072063}, and then transfers the original fundus image into polar coordinate system based on the detected disc center. Then the transferred image is fed into our M-Net, and generates the multi-label probability maps for OD and OC regions. Finally, the inverse polar transformation recovers the segmentation map back to the Cartesian coordinate.

\subsection{M-Net Architecture}

Our  M-Net  is an end-to-end multi-label deep network, which consists of four main parts. The first is a multi-scale layer used to construct an image pyramid input and achieve multi-level receptive field fusion. The second is a U-shape convolutional network, which is employed as the main body structure to learn a rich hierarchical representation. The third part is side-output layer that works on the early convolutional layers to support deep layer supervision.  Finally, a multi-label loss function is proposed to guarantee OD and OC segmentation jointly. 

\subsubsection{U-shape Convolutional Network}

In our paper, we modify the U-shape convolutional network (U-Net) in~\cite{Ronneberger2015} as the main body in our deep architecture. U-Net is an efficient fully convolutional neural network for the biomedical image segmentation.  Similar to the original U-Net architecture, our method consists of the encoder path (left side) and decoder path (right side). Each encoder path performs convolution layer with a filter bank to produce a set of encoder feature maps, and the element-wise rectified-linear non-linearity (ReLU) activation function is utilized. The decoder path also utilizes the convolution layer to output decoder feature map.  The skip connections transfer the corresponding feature map from encoder path and concatenate them to up-sampled decoder feature maps.

Finally, the high dimensional feature representation at the output of the final decoder layer is fed to a trainable multi-label classifier. In our method, the final classifier utilizes $1 \times 1$ convolutional layer with \textit{Sigmoid} activation as the pixel-wise classification to produce the probability map. For multi-label segmentation, the output is a $K$ channel  probability map, where $K$ is the class number ($K=2$ for OD and OC  in our work). The predicted probability  map corresponds to the class with maximum probability at each pixel. 

\subsubsection{Multi-scale Input Layer}

The multi-scale input or image pyramid has been demonstrated to improve the quality of segmentation effectively. Different from the other works, which fed the multi-scale images to multi-scream networks separately and combine the final output map in the last layer~\cite{Li2016TIP,Liu2017}, our M-Net employs the average pooling layer to downsample the image naturally and construct a multi-scale input in the encoder path. Our multi-scale input layer has following advantages: 1) integrating multi-scale inputs into the decoder layers to avoid the large  growth of parameters; 2) increasing the network width of decoder path. 

\subsubsection{Side-output Layer}

In our M-Net, we also introduce the side-output layer, which acts as a classifier that produces a companion local output map for early layers~\cite{Lee2015}. Let $\mathbf{W}$ denote the parameters of all the standard convolutional layers, and there are $M$ side-output layers in the network, where the corresponding weights are denoted as $\mathbf{w}=(\mathbf{w}^{(1)},...,\mathbf{w}^{(M)})$. The objective function of the side-output layer is given as:
\begin{equation}
\mathcal{L}_{s}(\mathbf{W}, \mathbf{w}) = \sum^M_{m=1} \alpha _m L^{(m)}_s(\mathbf{W}, \mathbf{w}^{(m)}),
\label{Eq_CNN_loss}
\end{equation}
where $\alpha _m$ is the loss function fusion-weight for each side-output layer ($\alpha _m = 0.25$ in our paper), $M$ is the side-output number, and  $L^{(m)}_s (,)$ denotes the multi-label loss of the  $m$-th side-output layer. To directly utilize side-output prediction map, we employ an average layer to combine all side-output maps as the final prediction map. The main advantages of the side-output layer are: first, the side-output layer back-propagates the side-output loss to the early layer in the decoder path with the final layer loss, which could relieve gradient vanishing problem and help the early layer training. It can be treated as a special bridge link between the loss and early layer; second, the multi-scale fusion has been demonstrated to achieve a high performance, and the side-output layer supervises the output map of each scale to output the better result.

\subsubsection{Multi-label Loss Function}
\label{sec-loss}

In our work, we formulate the OD and OC segmentation as a multi-label problem. The existing segmentation methods usually belong to the multi-class setting, which assign each instance to one unique label of multiple classes. By contrast, multi-label method learns an independent binary classifier for each class, and assigns each instance to multiple binary labels. Especially for OD and OC segmentation, the disc region overlays the cup pixels, which means the pixel marked as cup also has the label as disc. Moreover, for the glaucoma cases, the disc pixels excluded cup region shapes as a thin ring, which makes the disc label extremely imbalance to background label under the multi-class setting. Thus, multi-label method, considering OD and OC as two independent binary classifiers, is more suitable for addressing these issues. In our method, we propose a multi-label loss function based on Dice coefficient. The Dice coefficient is a measure of overlap widely used to assess segmentation performance when the ground truth is available~\cite{Crum2006}. Our multi-label loss function $L_s$ is defined as:
\begin{equation}
L_s = 1 - \sum_{k}^{K}  \dfrac{2 w_k \sum_{i}^{N}  p_{(k,i)} g_{(k,i)}}{\sum_{i}^{N}  p_{(k,i)}^2 + \sum_{i}^{N}  g_{(k,i)}^2} ,
\label{Eq_loss}
\end{equation}
where $N$ is the pixel number, $p_{(k,i)} \in [0, 1]$ and $g_{(k,i)} \in \{0, 1\} $ denote  predicted probability and binary ground truth label for class $k$, respectively.  $K$ is the class number, and $\sum_k w_k =1$ are the class weights. Our multi-label loss function in Eq.~(\ref{Eq_loss}) is equivalent to the traditional Dice coefficient by setting $K=1$. In our method, we set $K=2$ for OD and OC segmentation. Note that the Dice loss function indicates the foreground mask overlapping ratio, and can deals with the imbalance issue in the pixels of foreground (i.e., OD or OC) region and background. Under our multi-label setting, the pixel can be labeled as OD or/and OC independently. Thus, the imbalance issue does not exist between OD and OC.  $w_k$ in Eq.~(\ref{Eq_loss}) is the trade-off weight to control the contribution of OD and OC. For glaucoma screening, both the OD and OC are important, thus we set $w_k = 0.5$. Our multi-label loss function $L_s$ can be differentiated yielding the gradient as:
\begin{eqnarray}
\dfrac{\partial L_s}{\partial p_{(k,i)} }=  \sum_{k}^{K} 2 w_k  \left[ - \dfrac{ g_{(k,i)}  }{ \sum_{i}^{N}  p_{(k,i)}^2 + \sum_{i}^{N}  g_{(k,i)}^2 } \right. \nonumber \\
 \left. + \dfrac{ 2 p_{(k,i)} \sum_{i}^{N}  p_{(k,i)} g_{(k,i)} }{(\sum_{i}^{N}  p_{(k,i)}^2 + \sum_{i}^{N}  g_{(k,i)}^2)^2} \right] .
\end{eqnarray} 
This loss is efficiently integrated into back-propagation via standard stochastic gradient descent.

\subsection{Polar Transformation for Fundus Image}

\begin{figure}[!t]
	\begin{center}
		\includegraphics[width=1\linewidth]{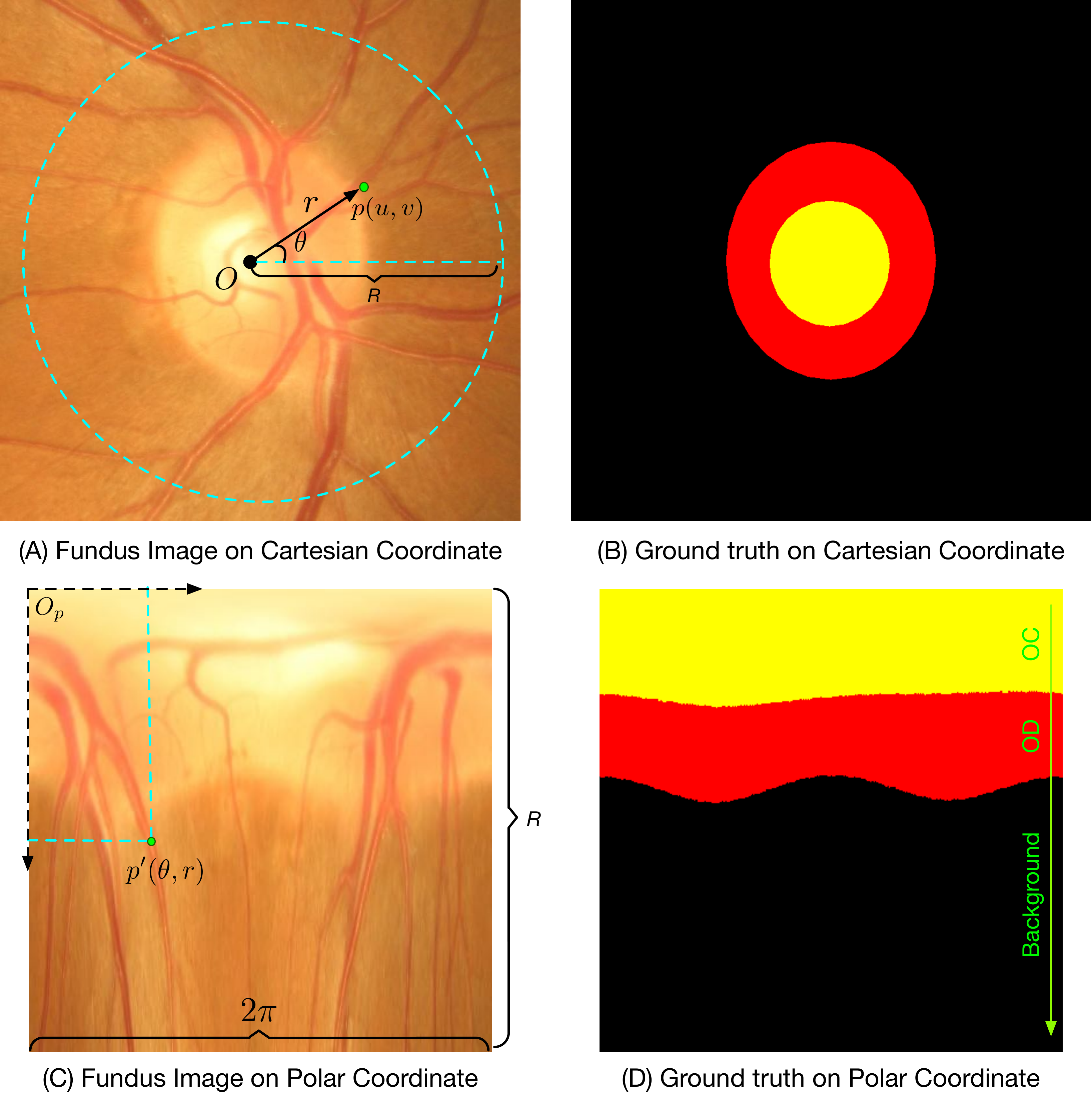}
		\caption{Illustration of the mapping from Cartesian coordinate system (A) to the polar coordinate system (C) by using the polar transformation. The point $p(u,v)$ in Cartesian coordinate corresponds to the point $p'(\theta, r)$ in polar coordinate. (B) and (D) are the corresponding ground truth, where yellow, red, and black regions denote the optic cup, optic disc, and background, respectively.}
		\label{img-polar}
	\end{center}
\end{figure}

In our method, we introduce a polar transformation for improving the OD and OC segmentation performance. The pixel-wise polar transformation transfers the original fundus image to the polar coordinate system. Let $p(u,v)$ denotes the point on fundus image plane, where the origin is set as the disc center $O(u_o, v_o)$, and $(u,v)$ is the Cartesian coordinates, as shown in Fig.~\ref{img-polar} (A). The corresponding point on polar coordinate system is $p'(\theta, r)$, as shown in Fig.~\ref{img-polar} (C), where $r$ and $\theta$ are the radius and directional angle of the original point $p$, respectively. The transform relation between the polar and Cartesian coordinates is as follow:
\begin{equation}
\left\{ {\begin{array}{{l}}
	u = r \cos \theta  \\
	v = r \sin \theta 
	\end{array}}   \right. 
\Leftrightarrow
\left\{ {\begin{array}{{l}}
	r=\sqrt{u^2 + v^2}  \\
	\theta = \tan ^{-1} v/u 
	\end{array}} . \right.
\label{Eq_pt}
\end{equation}
The height and width of transferred polar image are transformation radius $R$ and discretization $2\pi$.
The  polar transformation provides a pixel-wise representation of the original image in the polar coordinate system, which has the following properties:\\
\textbf{1) Spatial Constraint:} In the original fundus image, a useful geometry constraint is that the OC should be within the OD region, as shown in Fig.~\ref{img-polar} (B). But this redial relationship is difficult to implement in the original Cartesian coordinate. By contrast, our polar transformation transfers this redial relationship to a spatial relationship, where the regions of cup, disc, and background appear the ordered layer structure, as shown in Fig.~\ref{img-polar} (D). This layer-like spatial structure is convenient to use, especially some layer-based segmentation methods~\cite{Lang2013,Dufour2013} can be employed as the post-processing.\\
\textbf{2) Equivalent Augmentation:} Since the polar transformation is a pixel-wise mapping, the data augmentation on original fundus image is equivalent to that on polar coordinate. For example, moving the expansion center $O(u_o, v_o)$ is equivalent to the drift cropping transformations on polar coordinate. Using different transformation radius $R$ is same as augmenting with the various scaling factor. Thus the data augmentation for deep learning can be done during the polar transformation with various parameters.\\
\textbf{3) Balancing Cup Proportion:} In the original fundus image, the distribution of OC/background pixels is heavily biased. Even in the cropped ROI, the cup region still accounts for a low proportion. Using Fig.~\ref{img-polar} (B) as an example, the cup region only occupies about $4\%$. This extremely imbalance proportion easily leads the bias and overfitting in training the deep model. Our polar transformation flat the image based on OD center, that could enlarge the cup region by using interpolation and increase the OC proportion. As shown in Fig.~\ref{img-polar} (D), the ratio of cup region increases to $23.4\%$ over the ROI, which is more balanced than that in original fundus image. The balanced regions help avoid the overfitting during the model training and improve the segmentation performance further. 

Note that the method in~\cite{Chisako2009} also  utilizes the polar transformation to detect the cup outline based on the depth map estimated from stereo retinal fundus image pairs. Our work has significant difference compared to~\cite{Chisako2009}. 1) Motivations are different. The polar transformation used in~\cite{Chisako2009} aims at finding strongest depth edge in the radial direction as initial candidate points of cup border. In our work, we use polar transformation to obtain spatial constraint and augments the cup/disc region proportion. 2) Methods are different. The method in~\cite{Chisako2009} detects the OD and OC boundaries sequentially, and polar transformation only use for OC segmentation. Our method segments  OD and OC regions jointly, and considers their mutual relation under polar coordinate. 

\section{Experiments}
\label{sec_exp}

\subsection{Implementation}

Our M-Net is implemented with Python based on Keras with Tensorflow backend. During training, we employ stochastic gradient descent (SGD) for optimizing the deep model. We use a gradually decreasing learning rate starting from $0.0001$ and a momentum of $0.9$. 
\textcolor{red}{}
The transformation radius $R$ is set to $R=400$, and the directional angles are draw into $400$ distinct bins, thus the size of transferred polar image is $400\times400$.  The output of our M-Net is a 2-channel posterior probability map for OD and OC, where each pixel value represents the probability. A fixed threshold $0.5$ is employed to get a binary mask from the probability map. As same as that in previous works~\cite{Cheng2013,Cheng2017SVM}, the largest connected region in OD/OC mask is selected and the ellipse fitting is utilized to generate the final segmentation result.

\subsection{Segmentation Experiments}

We first evaluate the OD and OC segmentation performance.  We employ the ORIGA dataset~\cite{Zhang2010a} containing 650 fundus images with 168 glaucomatous eyes and 482 normal eyes. The 650 images with manual ground truth boundaries are divided into 325 training images (including 73 glaucoma cases) and 325 testing images (including 95 glaucoma) as same as that in~\cite{Xu2014,Cheng2017BOE}. To evaluate the segmentation performance, we use the overlapping error ($E$) and balanced accuracy ($A$) as the evaluation metrics for OD, OC, and rim regions:
\begin{equation}
E = 1 -  \dfrac{Area(S \bigcap G)}{Area(S \bigcup G)}, \; A =  \dfrac{1}{2} (Sen + Spe), 
\end{equation}
with
\begin{equation}
Sen =  \dfrac{TP}{TP+FN}, \;  Spe =  \dfrac{TN}{TN+FP}, 
\end{equation}
where $S$ and $G$ denote the segmented mask and the manual ground truth, respectively.  $TP$ and $TN$ denote the number of true positives and true negatives, respectively, and $FP$ and $FN$ denote the number of false positives and false negatives, respectively.
Moreover, we follow the clinical convention to compute the vertical cup to disc ratio (CDR) which is an important indicator for glaucoma screening. When CDR is greater than a threshold, it is glaucomatous, otherwise, healthy.  Thus an evaluation metric, named absolute CDR error $\delta_E$, is defined as: $\delta _E = | CDR_S - CDR_G |,$
where $CDR_G$ denotes the manual CDR from trained clinician, and $CDR_S$ is the CDR calculated by the segmented result.

\begin{figure*}
	\centering
	\subfigure[ORIGA dataset]
	{\includegraphics[width=.45\linewidth, height= .38\linewidth]{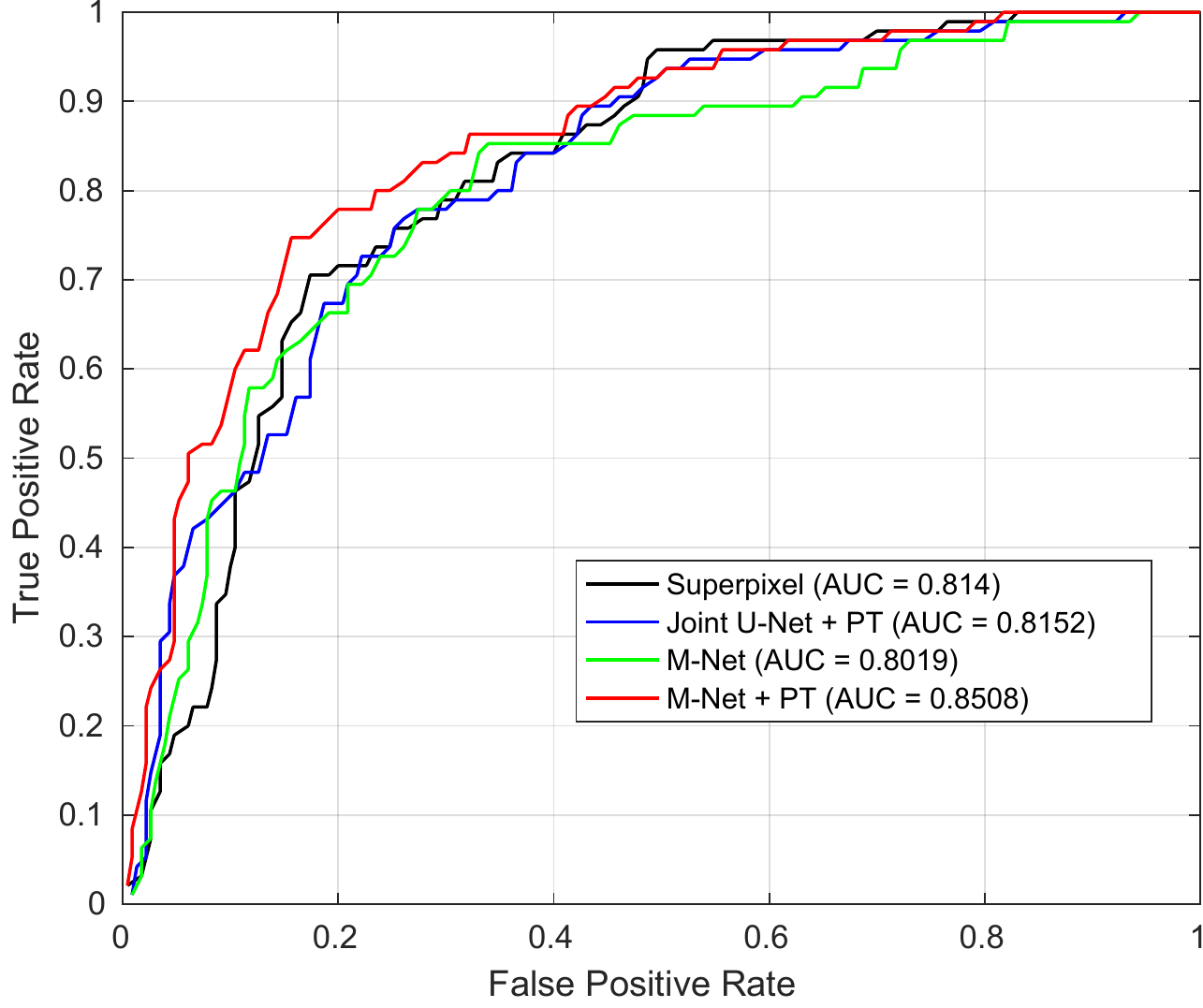}}  \;
	\subfigure[SCES dataset]
	{\includegraphics[width=.45\linewidth, height= .38\linewidth]{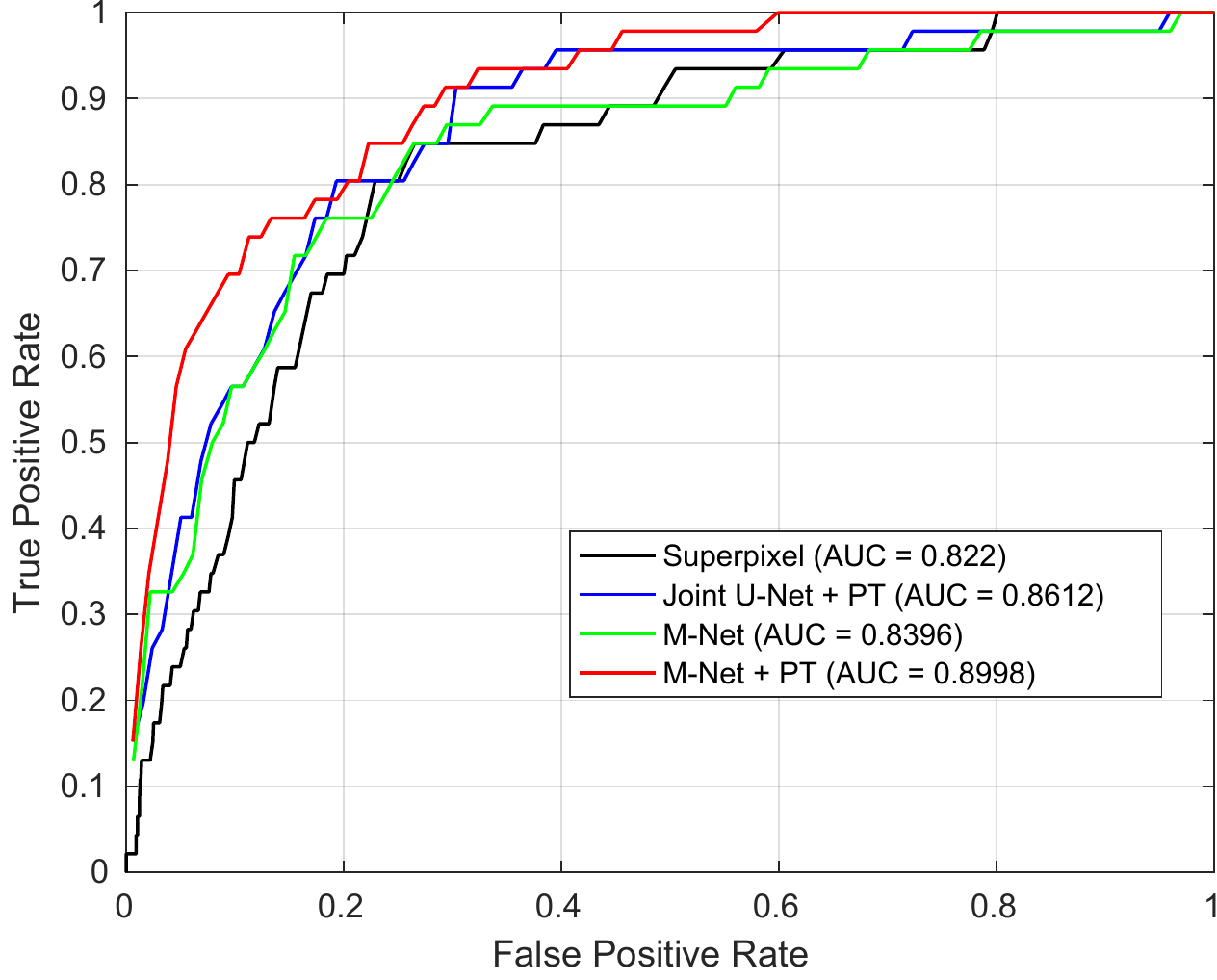}} \\ 
	\caption {The ROC curves with AUC scores for glaucoma screening based on the vertical cup to disc ratio (CDR) on (A) ORIGA and (B) SCES datasets. }
	\label{exp_auc}
\end{figure*}

\begin{table*}[!t]
	\centering
	\caption{Performance comparisons (\%) of the different methods on ORIGA Dataset. (PT: Polar Transformation.)}
	\begin{tabular}{|l||c|c|c|c|c|c|c|}
		\hline
		Method                       &   $E_{disc}$   &   $A_{disc}$   &   $E_{cup}$    &   $A_{cup}$    &  $E_{rim}$ &  $A_{rim}$ &  $\delta _E$   \\ \hline
		R-Bend~\cite{Joshi2011}      &     0.129      &       -        &     0.395      &       -        &             -              &             -              &     0.154      \\
		ASM~\cite{Yin2011}           &     0.148      &       -        &     0.313      &       -        &             -              &             -              &     0.107      \\
		Superpixel~\cite{Cheng2013}  &     0.102      &     0.964      &     0.264      &     0.918      &           0.299            &           0.905            &     0.077      \\
		LRR~\cite{Xu2014}            &       -        &       -        &     0.244      &       -        &             -              &             -              &     0.078      \\
		QDSVM~\cite{Cheng2017SVM}    &     0.110      &       -        &       -        &       -        &             -              &             -              &       -        \\
		U-Net~\cite{Ronneberger2015} &     0.115      &     0.959      &     0.287      &     0.901      &           0.303            &           0.921            &     0.102      \\ \hline
		Joint U-Net                  &     0.108      &     0.961      &     0.285      &     0.913      &           0.325            &           0.903            &     0.083      \\
		Our M-Net                    &     0.083      &     0.972      &     0.256      &     0.914      &           0.265            &           0.921            &     0.078      \\
		Joint U-Net + PT             &     0.072      &     0.979      &     0.251      &     0.914      &           0.250            &           0.935            &     0.074      \\
		Our M-Net + PT               & \textbf{0.071} & \textbf{0.983} & \textbf{0.230} & \textbf{0.930} &           \textbf{0.233}           &          \textbf{0.941}           & \textbf{0.071} \\ \hline
	\end{tabular} \\
	\label{Tab_Seg_score}%
\end{table*}%

We compare our M-Net with the several state-of-the-art OD/OC segmentation approaches: relevant-vessel bends (R-Bend) method in~\cite{Joshi2011}, active shape model (ASM) method in~\cite{Yin2011},  superpixel-based classification (Superpixel) method in~\cite{Cheng2013},  quadratic divergence regularized SVM  (QDSVM) method in~\cite{Cheng2017SVM}, and  low-rank superpixel representation (LRR) method in~\cite{Xu2014}. Additional, we compare  with the deep learning method U-Net~\cite{Ronneberger2015}. We report two results of U-Net, one is the original U-Net for segmenting OC and OD separately, and the other is U-Net utilized our multi-label loss function (Joint U-Net) for segmenting OC and OD jointly. We also provide segmentation results with/without the polar transformation (PT). The performances are shown in Table~\ref{Tab_Seg_score}.

R-Bend~\cite{Joshi2011} provides a parameterization technique based on vessel bends, and ASM~\cite{Yin2011} employs the circular Hough transform initialization to segment the OD and OC regions. These two bottom-up methods extract the OD and OC regions separately, and do not perform well on the ORIGA dataset.  Superpixel method in~\cite{Cheng2013} utilizes superpixel classification to  detect the OD and OC boundaries. It obtains a better performance than the other two bottom-up methods~\cite{Joshi2011,Yin2011}. The methods in LRR~\cite{Xu2014} and QDSVM~\cite{Cheng2017SVM} obtain good results.  However, they only focus on either OD or OC segmentation, and can not calculate the CDR for glaucoma screening. Joint U-Net with our multi-label loss utilizes the mutual relation of OD and OC, and obtains a better performance than that in traditional U-Net~\cite{Ronneberger2015}.
Our M-Net with multi-scale input and side-output layers achieves a higher score than single-scale network  and superpixel method~\cite{Cheng2013}. It demonstrates that the  multi-scale input and side-output layers are useful to guide the early layer training. 

The polar transformation as one contribution of our work augments the proportion of cup region. One major advantage is that the polar transformation augments the proportion of cup region, and makes the area of the disc/cup and background more balance. The balanced regions help avoid the overfitting during the model training and improve the segmentation performance further.  From Table~\ref{Tab_Seg_score}, polar transformation reduces about $0.03$ in Joint U-Net and $0.02$ in M-Net on $E_{cup}$ scores. Note that the performance of Joint U-Net with PT is slight better than that in M-Net without PT. It shows that the gains of the polar transformation may be higher than that using multi-scale input and side-output layers. Finally, our M-Net with PT achieves the best performance, and outperforms other state-of-the-art methods. 

\begin{figure*}[!t]
	\begin{center}
		\includegraphics[width=1\linewidth]{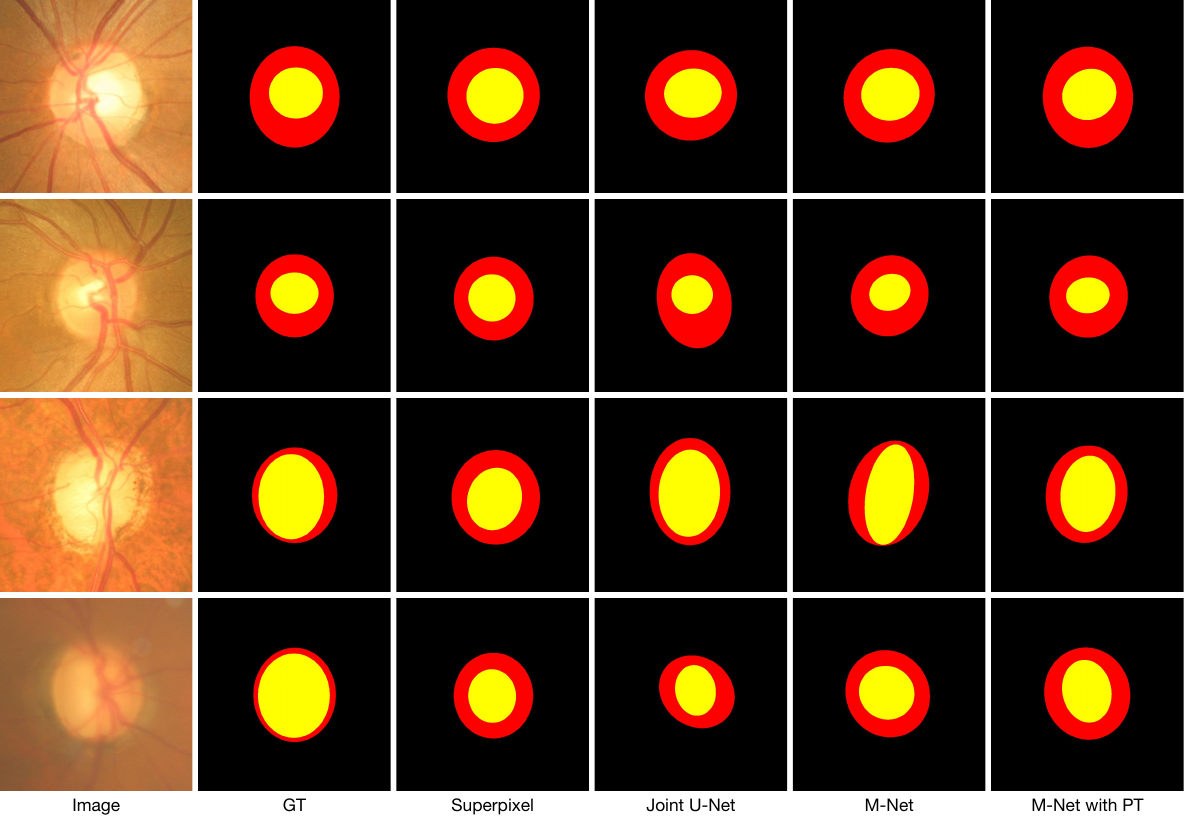}
		\caption{The visual examples of optic disc and cup segmentation, where the yellow and red region denote the cup and disc segmentations, respectively. From the left to right: fundus image, ground truth (GT), Joint U-Net, our M-Net and M-Net with polar transformation (PT). The last row shows the failed case.}
		\label{img-exp-seg}
	\end{center}
\end{figure*}

Fig.~\ref{img-exp-seg} shows the visual examples of the segmentation results, where the first two rows are normal eyes and the rest rows are glaucoma cases. For the superpixel method~\cite{Cheng2013}, the segmented OC is smaller than ground truth in glaucoma cases, which may cause an under-estimated CDR. The deep learning methods (e.g., Joint U-Net and M-Net) obtain more accurate cup boundary, but it easily generates a larger OD. By contrast, our M-Net with PT can effectively and accurately segment OD and OC regions. The last row in Fig.~\ref{img-exp-seg} shows a challenging case for segmentation, where the image is blurred and has low-contrast for identifying the OC boundary. For this case, all the methods fail to produce accurate OC segmentation. This issue could potentially be addressed in future work through the use of more powerful network or additional image enhancement pre-processing.

\subsection{Glaucoma Screening}

We also evaluate the proposed method on glaucoma screening by using the calculated CDR value. Two datasets are used, one is ORIGA dataset, the second is Singapore Chinese Eye Study  (SCES) dataset. For ORIGA dataset, we employ 325 image for training and rest for testing, as same as used in segmentation experiment. For the SCES dataset, it consists of 1676 images, of which 46 ($\sim 3\%$) are glaucoma cases. Since the SCES dataset provides only clinical diagnoses, it will be used only to assess the diagnostic performance of our system. We use all the 650 images in ORIGA dataset for training and the whole 1676 images of SCES for testing. We report the Receiver Operating Characteristic (ROC) curve and area under ROC curve (AUC) as the overall measure of the diagnostic strength. The  performances  for glaucoma screening based on CDR are shown in Fig.~\ref{exp_auc}. 

From the glaucoma screening results, we have the following observations: 1) The non-deep learning method, superpixel~\cite{Cheng2013}, produces a competitive performance ($AUC=0.814$) on ORIGA dataset, which is better than M-Net ($AUC=0.8014$). But its performance is lower than others on SCES dataset. 2)  The Joint U-Net with PT obtains   higher scores than superpixel~\cite{Cheng2013} and U-Net on both ORIGA and SCES datasets. 3) Our M-Net with PT achieves the best performances on ORIGA dataset ($AUC=0.8508$) and SCES dataset ($AUC=0.8997$). Especially, our M-Net with PT has more than $5 \%$ improvement on AUC than M-Net without PT, which demonstrates the effectiveness of polar transformation on glaucoma screening. 4)  Our method also outperforms other deep learning based diagnostic method. For example, the deep learning method in~\cite{Chen2015MICCAI} provides a glaucoma screening system by using deep visual feature, which obtained $AUC = 0.838$ and $AUC = 0.898$ on ORIGA and SCES dataset. However, it can not provide the CDR value as a clinical explanation. Our result of M-Net with PT is comparable to that of deep system~\cite{Chen2015MICCAI}. 5) Final, all the deep learning based methods have better performances in SCES dataset than those in ORIGA dataset. One possible reason is that the size of training set on ORIGA is only 325. The more training data promotes the representation capability of deep learning.
  
\subsection{Discussion}

\subsubsection{Running Time}

The entire training phase of our method takes about 5 hours on a single NVIDIA Titan X GPU (100 iterations). However, the training phase could be done offline. In online testing, it costs only $0.5 s$ to generate the final segmentation map for one fundus image, which is faster than the existing methods, e.g., superpixel method~\cite{Cheng2013} takes $10 s$, ASM method~\cite{Yin2011} takes $4 s$, R-Bend method~\cite{Joshi2011} takes $4 s$, sequential segmentation of OD and OC by using original U-Net~\cite{Ronneberger2015} takes $1s$. 

\subsubsection{Repeatability Experiment }

\begin{table}[!t]
	\centering
	\caption{Correlation Coefficients of Repeatability Experiment. }
	\begin{tabular}{|l||c|c|c|}
		\hline
		Data & Glaucoma (n=39) & Normal (n=1481) &  All (n=1520)  \\ \hline
		Coefficient &  0.8833  & 0.8262 & 0.8357 \\
		$p$-value &  $< 0.0001$  & $< 0.0001$ & $< 0.0001$ \\ \hline
	\end{tabular} \\
	\label{Tab_repeat}%
\end{table}%

\begin{figure}[!t]
	\begin{center}
		\includegraphics[width=0.9\linewidth]{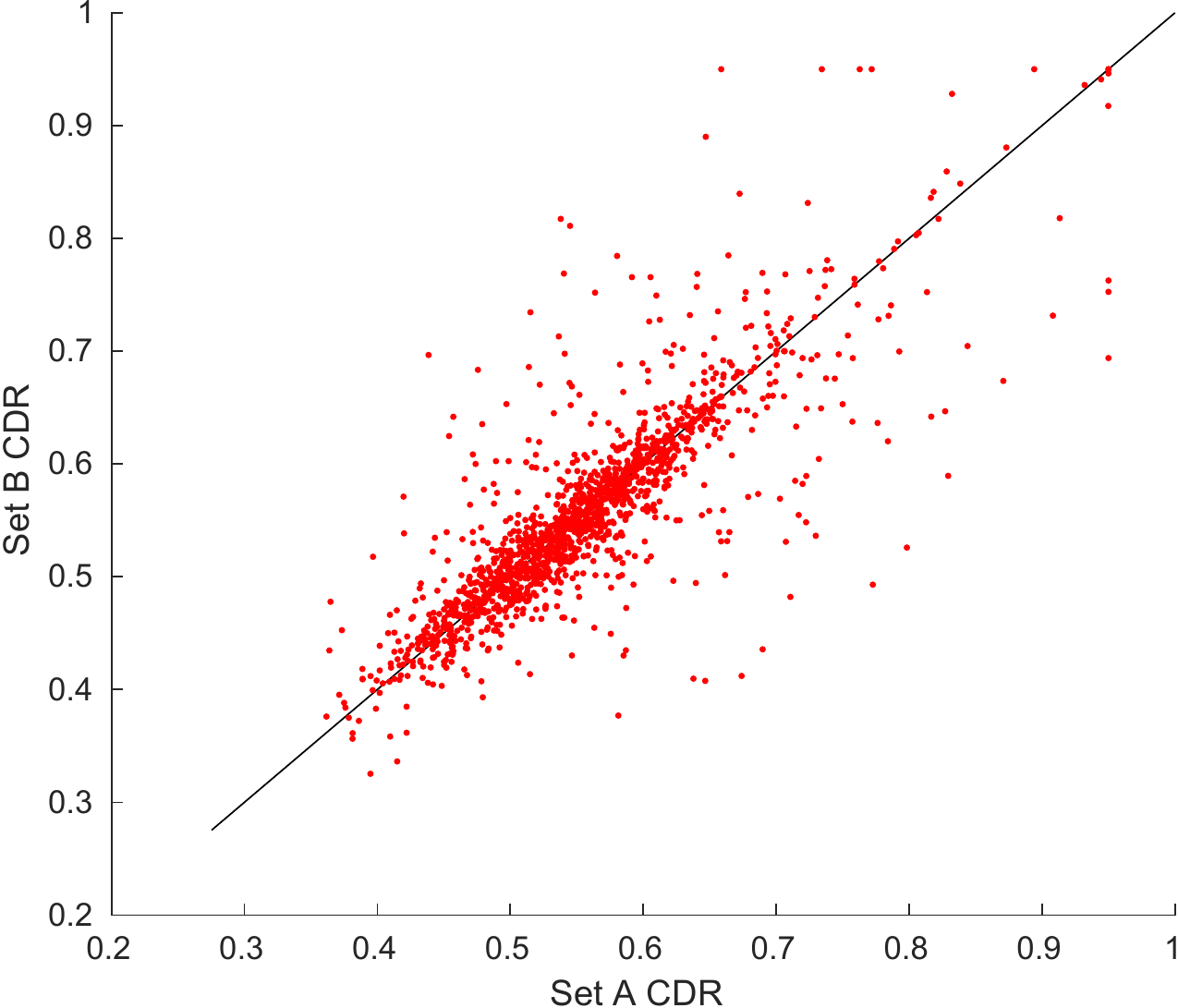}
		\caption{Scatter plot of the CDR correspondence on the repeatability dataset.}
		\label{img-repeat}
	\end{center}
\end{figure}

In this experiment, we evaluate the repeatability of proposed method. We collect a repeatability dataset with two corresponding sets (A and B) consisting of 1520 fundus image pairs. For each pair, one image is selected from SCES dataset, and other one is the different image from the same visit. We run our proposed method on these two sets, and calculate the correlation coefficients of CDR values. Table~\ref{Tab_repeat} reports the repeatability test result, and the scatter plot of the CDR correspondence is shown in Fig.~\ref{img-repeat}. As it can be seen, our method gets $p< 0.0001$ of $p$-value and appears a good repeatability.

\subsubsection{Clinical Measurement}

Our M-Net method segments the whole OD and OC regions, which could be used to calculate other clinical measurements. In this experiment, we evaluate the rim to disc area ratio (RDAR)~\cite{Jonas362} defined as: 
${Area(Disc - Cup)}/{Area(Disc)}$.
The comparison of CDR and  RDAR is shown in Table~\ref{Tab_RDAR}, where our M-Net with PT obtains the best screening performance based on RDAR value on both datasets, which is consistent with the experiment based on CDR. Moreover, CDR measurement shows a better screening performance than RDAR. The possible reasons may be that the rim is calculated by subtracting of disc and cup regions, which contains the errors of both disc and cup segmentations. Thus the rim error is larger than cup error. This also observes from Table~\ref{Tab_Seg_score}, where the rim error ($E_{rim} = 0.233$) is larger than cup error ($E_{cup} = 0.230$) based on our M-Net with PT. Moreover, since the central retinal vessel trunk usually locates in the nasal optic disc sector~\cite{Jonas12322002}, it renders difficult the automatic delineation of the optic disc region in horizontal. Thus, the vertical disc and cup may obtain the higher accuracy than that in horizontal.

\begin{table}[!t]
	\centering
	\caption{AUC performance  of the different Clinical Measurements. (CDR: vertical cup to disc ratio. RDAR: rim to disc area ratio)}
	\begin{tabular}{|l||c|c|c|c|}
		\hline
		& \multicolumn{2}{c|}{ORIGA Dataset} & \multicolumn{2}{c|}{SCES Dataset} \\ \hline
		Method           & $CDR$  &          $RDAR$           & $CDR$  &          $RDAR$          \\ \hline
		Our M-Net        & 0.8019 &          0.7981           & 0.8397 &          0.8290          \\
		Joint U-Net + PT & 0.8152 &          0.7921           & 0.8612 &          0.8003          \\
		Our M-Net  + PT  & 0.8508 &          0.8425           & 0.8998 &          0.8488          \\ \hline
	\end{tabular} \\
	\label{Tab_RDAR}%
\end{table}%

\section{Conclusion}
\label{sec_conclusion}

In this paper, we have developed a deep learning architecture, named M-Net, which solves the OD and OC segmentation jointly into a one-stage multi-label framework. The proposed  M-Net employed the U-shape convolutional network as the body structure. And the multi-scale layer constructed an image pyramid to fed a multi-level inputs, while the side-output Layers acted as a early classifier to produce the companion local prediction maps for early scale layers. A multi-label loss function has been proposed to guarantee the final output for segmenting OD and OC together. For improving the segmentation result further, we also have introduced a polar transformation to transfer the  original fundus image to the polar coordinate system. We have demonstrated that our system produces state-of-the-art segmentation results on ORIGA dataset. Simultaneously, the proposed method also obtained the satisfactory glaucoma screening performances by using the calculated CDR on both ORIGA and SCES datasets. The work implementation details are available at \url{http://hzfu.github.io/proj_glaucoma_fundus.html}.

\ifCLASSOPTIONcaptionsoff
\newpage
\fi

{
\bibliographystyle{IEEEtran}
\bibliography{IEEEabrv,Deep_CDR}
}

\end{document}